\documentclass[final]{cvpr}

\usepackage{times}
\usepackage{epsfig}
\usepackage{graphicx}
\usepackage{amsmath}
\usepackage{amssymb}
\usepackage{smil}
\usepackage{multirow}
\usepackage{algorithm}
\usepackage{algorithmic}

\def\fullname{Anchor-Captioner}
\def\shortname{A-Cap}

\def\mata{\textcolor{black}}

\def\wzy{\textcolor{black}}
\def\cph{\textcolor{black}}
\def\rev{\textcolor{black}}

\usepackage[pagebackref=false,breaklinks=true,colorlinks,bookmarks=false]{hyperref}

\def\cvprPaperID{1215} 
\def\confYear{CVPR 2021}

\begin{document}

\title{Towards Accurate Text-based Image Captioning with \\ Content Diversity Exploration}

\author{
    Guanghui Xu$^{1,2}$\thanks{Authors contributed equally.},
    Shuaicheng Niu$^{1*}$, Mingkui Tan$^{1,4}$, Yucheng Luo$^{1}$, Qing Du$^{1,4}$\thanks{Corresponding author}, Qi Wu$^{3}$ \\
    $^{1}$South China University of Technology,
    $^{2}$Pazhou Laboratory, $^{3}$ University of Adelaide \\
    $^{4}$Key Laboratory of Big Data and Intelligent Robot, Ministry of Education, \\
    {\tt\small{sexuguanghui}@mail.scut.edu.cn},
    {\tt\small \{mingkuitan, duqing\}@scut.edu.cn}, {\tt\small{qi.wu01@adelaide.edu.au}}
}

\maketitle
\begin{abstract}
Text-based image captioning (TextCap) which aims to read and reason images with texts is crucial for a machine to understand a detailed and complex scene environment, considering that texts are omnipresent in daily life. This task, however, is very challenging because an image often contains complex texts and visual information that is hard to be described comprehensively. Existing methods attempt to extend the traditional image captioning methods to solve this task, which focus on describing the overall scene of images by one global caption. This is infeasible because the complex text and visual information cannot be described well within one caption. To resolve this difficulty, we seek to generate multiple captions that accurately describe different parts of an image in detail. To achieve this purpose, there are three key challenges: 1) it is hard to decide which parts of the texts of images to copy or paraphrase; 2) it is non-trivial to capture the complex relationship between diverse texts in an image; 3) how to generate multiple captions with diverse content is still an open problem. To conquer these, we propose a novel Anchor-Captioner method. Specifically, we first find the important tokens which are supposed to be paid more attention to and consider them as anchors. Then, for each chosen anchor, we group its relevant texts to construct the corresponding anchor-centred graph (ACG). Last, based on different ACGs, we conduct the multi-view caption generation to improve the content diversity of generated captions. Experimental results show that our method not only achieves SOTA performance but also generates diverse captions to describe images.
\end{abstract}

\section{Introduction}

The texts are omnipresent in \wzy{our} daily life and play an important role in helping humans or intelligent robots to understand the physical world~\cite{Fine2014SensorySD}.
In the image captioning area, the texts contained in images are also of critical importance and often provide valuable information~\cite{Biten2019SceneTV,hu2019iterative,huang2020graph,Mishra2019OCRVQAVQ,singh2019towards} for caption generation.
In this sense,
Sidorov~\etal~\cite{Sidorov2020TextCapsAD} propose a fine-grained image captioning task, \ie~text-based image captioning (TextCap), aiming to generate image captions that not only `describe' visual contents but also `read' the texts in images, such as billboards, road signs, commodity prices and etc. 
\wzy{This task is very practical since the fine-grained image captions with rich text information can aid visually impaired people to comprehensively understand their surroundings~\cite{Fine2014SensorySD}}

Some preliminary tries for the TextCap task seek to directly extend existing image captioning methods~\cite{Anderson2018BottomUpAT,hu2019iterative,Huang2019AttentionOA} to this new setting.
\wzy{However, such methods usually tend to describe prominent visual objects or overall scenes without considering the texts in images.}
\cph{Recently, M4C-Captioner~\cite{Sidorov2020TextCapsAD} tries to use additional OCR tools~\cite{Baek2019CharacterRA,borisyuk2018rosetta,Liu2020ABCNetRS} to recognise texts in images. It is still hard to well describe the complex text and visual information within one caption. 
To resolve this difficulty, we propose to generate multiple diverse captions focusing on describing different parts of an image. However, there are still some challenges.}

\begin{figure}[t]
    \centering
    \includegraphics[width=0.45\textwidth]{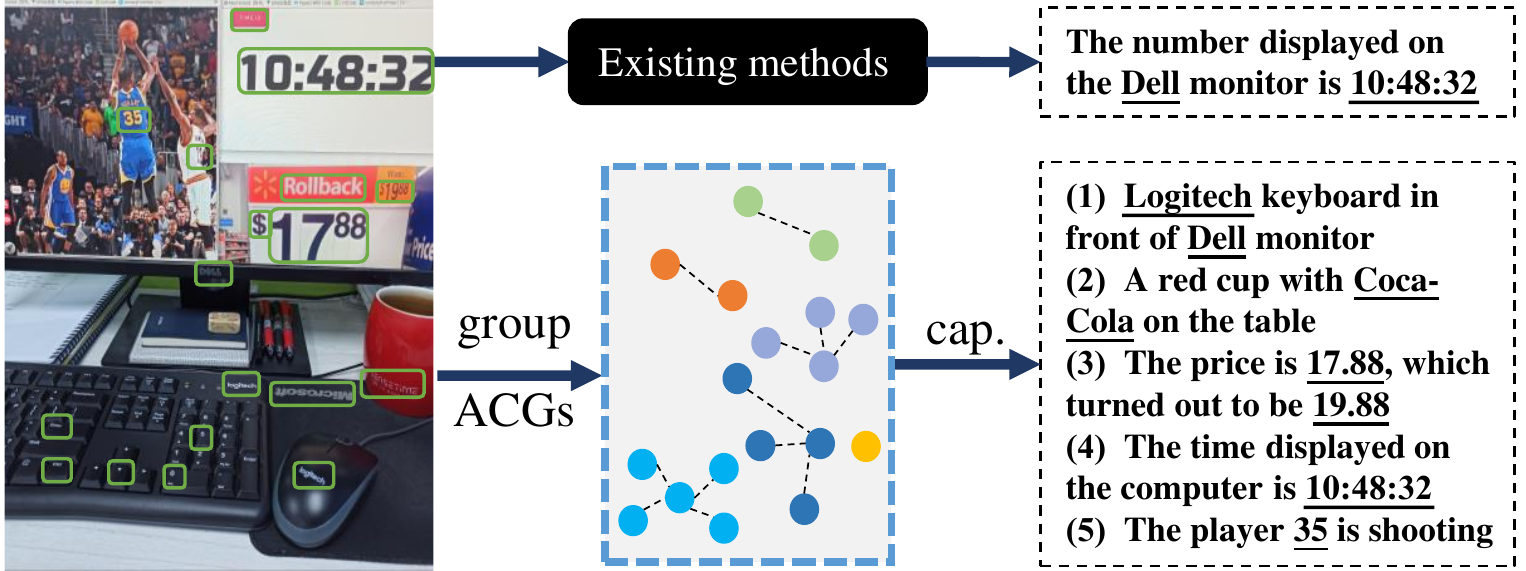}
    \caption{Comparison with existing methods. For a given image, existing methods tend to generate only one global caption. Unlike them, we first select and group texts to anchor-centred graphs (ACGs), and then decide \wzy{which} parts of the texts to copy or paraphrase. Our method is able to achieve higher accuracy and generate diverse captions to describe the image from different views.}
    \label{fig:acgs_captioning}
\end{figure}

First, \wzy{it is hard to decide which parts of the texts are most crucial for describing the images}. As shown in Figure~\ref{fig:acgs_captioning}, an image often contains a lot of texts, but only a small part of the texts play a key role in caption generation. For example, \emph{a PC keyboard contains many letters}, but we do not need a caption that covers all the recognised letters.

Second, it is non-trivial to capture the complex relationship between diverse texts in an image. The correct understanding of such a relationship is essential for \wzy{generating accurate captions}. For example, \emph{to accurately describe a cup, we might use its brand and capacity}. But these texts have no relevance to the contents on the computer screen. 

More critically, how to generate multiple captions describing different contents remains unknown. Current image captioning methods~\cite{Anderson2018BottomUpAT,hu2019iterative,Huang2019AttentionOA} often only generate a content-monotone caption. They tend to focus on a small part of the contents in the image, \emph{such as the time in the monitor in Figure~\ref{fig:acgs_captioning}}. To comprehensively describe an image, one better solution is to generate diverse captions, where each caption focuses on describing one relevant part. 

To address the above issues, we design a new \fullname{} architecture that mainly consists of two key modules,~\ie~an anchor proposal module (AnPM) and an anchor captioning module (AnCM). 
Specifically, AnPM is proposed to understand the texts in an image and to capture the complex relationships between \wzy{different texts}.
To be specific, we first employ an anchor predictor to rank the importance of each token. Then, we choose several important tokens to decide \wzy{which parts of texts are most informative and need to be carefully considered}. After that, considering each chosen token as an anchor, we use a recurrent neural network to model its complex relationships with other tokens and to construct an anchor-centred graph (ACG) for each anchor. Each ACG denotes a group of the relevant tokens which are supposed to be included in the same caption. 
Based on the different ACGs for an image, we apply AnCM to generate diverse captions that cover various OCR tokens. To be specific, we first generate a visual-specific caption to model global visual information. 
\wzy{Then, we take each ACG as guidance to refine the visual caption and generate multiple text-specific captions that contain fine-grained text information.}
Extensive experimental results \cph{on TextCaps dataset} demonstrate the effectiveness of our proposed method.

In summary, our main contributions are as follows:
\begin{enumerate}
    \item \wzy{We propose to exploit fine-grained texts information to generate multiple captions that describe different parts of images}, instead of generating a single caption to handle them as a whole.

    \item We propose an anchor proposal module (AnPM) and an anchor captioning module (AnCM) to select and group texts to anchor-centred graphs (ACGs) and then generate diverse captions based on ACGs.
    
    \item We achieve the state-of-the-art results on TextCaps dataset, in terms of both accuracy and diversity. 
\end{enumerate}

\section{Related work}
\noindent\textbf{Image captioning} aims to automatically generate textual descriptions of an image, which is an important and complex problem since it combines two major artificial intelligence fields: natural language processing and computer vision. 
Most image captioning models~\cite{Anderson2018BottomUpAT,Gan2017SemanticCN,Vaswani2017AttentionIA,Vinyals2015ShowAT,MMA2020Wang,Xu2015ShowAA} use CNNs to encode visual features and apply RNNs as language decoder to generate descriptions. 
Some works~\cite{Ghazvininejad2019MaskPredictPD,Lee2018DeterministicNN,Liu2021HowTT,NIPS2017_6775} propose to further refine the generated sentences with multiple decoding passes. \rev{NBT~\cite{Lu2018NeuralBT} first generates a template without specifics and then fills it with ‘object’ words.}
\rev{RL-based methods}~\cite{guo2019nat,Liu2017ImprovedIC,Niu2020DisturbanceimmuneWS,Rennie2017SelfCriticalST} model the sequence generation as Markov Decision Process~\cite{williams1992simple} and directly maximise the metric scores.

To generate diverse image captions, many works try to control the generation in terms of style and contents.  
The style controllable methods~\cite{Gan2017StyleNetGA,Guo2019MSCapMI,Mathews2018SemStyleLT} usually require additional annotations for training, such as a pair of labelled captions with different styles.
Other parallel works focus on controlling the contents of the generated captions. Johnson~\etal~\cite{Johnson2016DenseCapFC} are the first to propose the dense captioning task to describe the visual objects in a sub-region~\cite{zeng2020dense}. 
\rev{Signal-based methods}~\cite{Chen2020SayAY,Cornia2019ShowCA,Deng2020LengthControllableIC,Deshpande2019FastDA} sample different predictions based on the control signals to obtain diverse image captions. 
Our work can be seen as text-based dense captioning and \rev{aims to generate multi-view captions}.
\begin{figure*}[t]
    \centering
    \includegraphics[width=\textwidth]{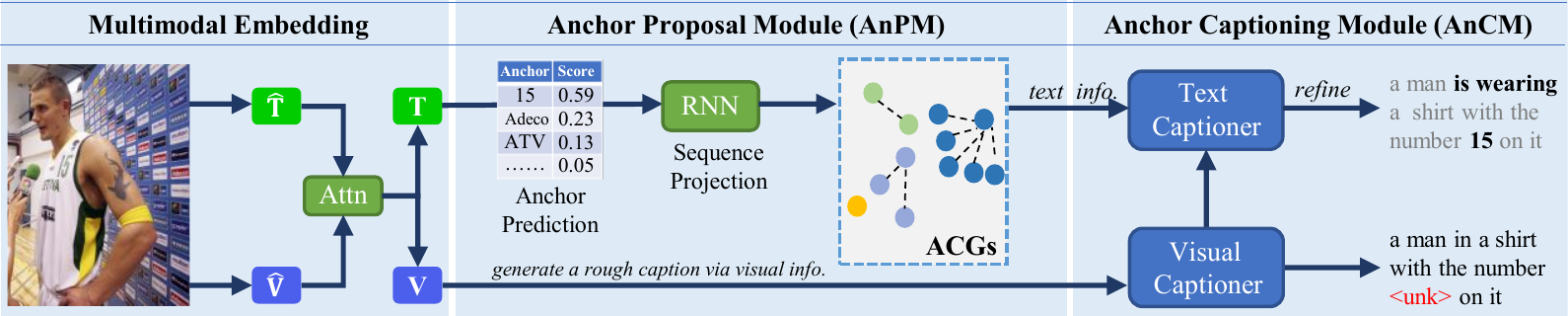}
    \caption{An illustration of \fullname{}. Given an input image, (1) we first extract text and visual features $(\widehat{\bT},\widehat{\bV})$ independently. Then, we fuse them to obtain multimodal features ($\bT, \bV$) via self-attention; 
    (2) Following that, AnPM chooses a series of anchor-tokens based on the anchor predictions and then groups the relevant tokens by constructing anchor-centred graphs (ACGs).
    (3) Lastly, AnCM employs a visual-captioner to output a global visual-specific caption, and then uses a text-captioner to generate multiple text-specific captions based on the above global caption and ACGs.
    In this figure, we only show the generated caption with the ACG of the top-1 score.}
    \label{fig:overview}
\end{figure*}

\noindent\textbf{Text-based image captioning} aims to generate captions describing both the visual objects and written texts. 
Intuitively, the text information is important for us to understand the image contents. 
However, the existing image captioning datasets~\cite{Krishna2016VisualGC,Lin2014MicrosoftCC} have a bias that only \wzy{describes} the salient visual objects in the image while ignoring the text information. 
As a result, most image captioning models~\cite{Anderson2018BottomUpAT,Gan2017SemanticCN,Vaswani2017AttentionIA,Vinyals2015ShowAT,Xu2015ShowAA} unable to `read' the texts since they don't pay attention to improve such ability. 
In this sense, Sidorov~\etal~\cite{Sidorov2020TextCapsAD} introduce a novel dataset, namely TextCaps, which requires a captioning model not only to `watch' visual contents but also `read' the texts in images. 
They introduce a benchmark M4C-Captioner~\cite{Sidorov2020TextCapsAD}, which is simply extended from the M4C~\cite{hu2019iterative} (for TextVQA). Specifically, they feed all the detected texts and visual contents into their captioning model to generate a global caption for an input image.
\wzy{However}, it is difficult for a single caption to cover all the \wzy{multimodal} information, and the overlooked parts may be the information that people are interested in.

Different from existing methods, we propose an anchor proposal module to understand the relationship within OCR tokens and group them to construct anchor-centred graphs (ACGs). With the help of ACGs, our method is able to better describe the input image by generating diverse captions.

\section{Proposed method}
We study text-based image captioning (TextCap) which aims to read and reason an image with texts to generate detailed captions. 
This task is very challenging because it is difficult to comprehensively describe images with rich information. Existing methods tend to generate one global caption that tries to describe complex contents in an image. However, such methods are unfeasible when the image contains a large number of redundant visual objects and diverse texts. 
To accurately describe an image, one better solution is to generate multiple captions from different views. However, several challenges still exist.
First, it is hard to decide \wzy{which} parts of the texts of images to copy or paraphrase when images contain a lot of texts. Second, it is non-trivial to exploit the correct relationship between diverse texts in an image, which, however, is essential to accurately describe the image. More critically, how to generate multiple captions from different views for comprehensively describing images is still unknown.

\wzy{
In this paper, we propose a new captioning method \fullname{} that aims to accurately describe images \cph{by using} content diversity exploration.
As shown in Figure~\ref{fig:overview}, \fullname{} has two main components,~\ie~an anchor proposal module (AnPM) and an anchor captioning module (AnCM). 
AnPM chooses important texts as anchors and constructs anchor-centred graphs (ACGs) to model complex relationship between texts.
AnCM takes different ACGs as input to generate multiple captions that describe different parts of an image.
In this way, our method is able to choose important relevant texts to describe and also has the ability to generate diverse captions for comprehensively understanding images.
}

\subsection{Multimodal embedding}
To generate captions for an image, we first use a pre-trained Faster RCNN~\cite{ren2017faster} model to extract $N$ visual objects and recognise $M$ OCR tokens by the Rosetta OCR~\cite{borisyuk2018rosetta}. 

\noindent\textbf{Visual embedding.} For the $i$-th visual object, the Faster RCNN model outputs appearance feature $\bv_i^{a} \in \mmR^{d}$ and a 4-dimensional bounding box coordinate $\bv_i^{b}$. To enrich the visual representation, we apply a linear layer $f_1$ with LayerNorm~\cite{layernormLei} to project {the} above features as $\widehat{\bv}_i = f_1 ([\bv_i^{a}, \bv_i^{b}]),$
where $[\cdot,\cdot]$ is a concatenation operation.

\noindent\textbf{Token embedding.} For each recognised OCR token, we also use its appearance feature $\bt_i^{a}$ and bounding box coordinate $\bt_i^{b}$. Apart from these features, following the M4C-Captioner~\cite{Sidorov2020TextCapsAD}, we adopt two additional text features to further \wzy{enrich} the representations, including FastText feature $\bt^f_{i}$ and PHOC (pyramidal histogram of characters) feature $\bt^p_{i}$. In particular, $\bt^f_{i}$ is a pretrained word-level embedding for written texts while $\bt^p_{i}$ is a character-level embedding for capturing what characters are present in the tokens. 
Based on the rich representations of OCR tokens, we calculate OCR token features by $\widehat{\bt}_i = f_2 ([\bt_i^a, \ \bt_i^b, \ \bt_i^f, \ \bt_i^p]),$
where $f_2$ is a linear layer with LayerNorm to ensure that token embedding has the same scale as visual embedding.

\noindent\textbf{Multimodal embedding fusion.} Based on the above, we obtain visual embedding $\widehat{\bV} \small{=} [\widehat{\bv}_1,...,\widehat{\bv}_N]^\top$ and token embedding $\widehat{\bT} \small{=} [\widehat{\bt}_1,...,\widehat{\bt}_M]^\top$. 
Since both the OCR tokens and visual objects are visual contents and exist in images as a whole, {it is necessary to model their interaction}. Formally, given $\widehat{\bV}$ and $\widehat{\bT}$, we use an $L_1$-layer Transformer module $\Psi (\cdot; \theta_{a})$ to {obtain more informative features} via \wzy{a self-attention operation} as
\begin{align}
    \bV, \bT = \Psi \ ([\widehat{\bV}, \widehat{\bT}]; \theta_{a}).
\end{align}

\subsection{Anchor proposal module}
\label{sec:apm}
Based on the multimodal embeddings ($\bV, \bT$), existing TextCap methods such as M4C-Captioner~\cite{Sidorov2020TextCapsAD} simply treat the texts in an image as another kind of visual information and feed them to a captioning module without distinction. However, compared with ambiguous visual information, the texts in images are essential to describe images and thus need to be considered carefully.

To this end, we propose the anchor proposal module (AnPM) to determine which OCR \wzy{tokens} should be paid more attention to describe. 
Inspired by the region proposal network (RPN), AnPM first performs anchor prediction among OCR tokens to output a score for each token and choose a series of important tokens as anchors. 
After that, to model the complex relationship between tokens, AnPM groups relevant tokens to construct the corresponding anchor-centred graph (ACG) for each anchor.
Now, we introduce how to construct ACGs in detail.


\noindent\textbf{Anchor prediction.} 
Intuitively, different OCR tokens play different roles in caption generation. However, it is hard to decide which texts should be paid more attention to.
To this end, based on the text features $\bT$, we apply a linear layer $\phi$ as an anchor predictor to predict a score for each token as
\begin{align}
    \label{eq:anchor_select}
    \bs_{anchor} &= \mathrm{Softmax} \ (\phi(\bT)).
\end{align}

In the anchor score $\bs_{anchor} \small{\in} \mmR^{M}$, each element indicates the importance weight of each OCR token.
In training, we adopt the OCR token with the highest score as the anchor by argmax operation, denoted as 
\begin{align}
    \bT_{anchor} = \bT_{i,:}, ~~~\mathrm{where} \ i = \mathrm{argmax} (\bs_{anchor}).
\end{align}
After that, we can obtain the anchor embedding $\bT_{anchor}$.
{During the inference phase, we choose OCR tokens with top-$K$ scores as anchors.}

\noindent\textbf{Anchor-centred graph construction.} 
In this paper, we employ a RNN module and take $\bT_{anchor}$ as the initial hidden state to model the potential dependence between the anchor and different tokens.
The ACG construction for the anchor $\bT_{anchor}$ can be formulated as:
\begin{equation}
\begin{split}
    \bT_{graph} &= \mathrm{RNN} \ (\bT, \ \bT_{anchor}), \\
    \label{eq:graph_cons}
    \bs_{graph} &= \sigma(f_{3}(\bT_{graph})),
\end{split}
\end{equation}
where $\bT_{graph} \in \mmR^{M \times d}$ denotes the updated token feature and $f_{3}$ is a linear layer followed by Sigmoid activation function $\sigma$ to output the graph score $\bs_{graph} \in \mmR^{M}$ for all $M$ tokens. 
After that, we concatenate $\bT_{anchor}$ and its relevant tokens to construct the ACG $\mG$ as follows: 
\begin{align}
    \label{construct_graph}
    \mG = & [\bT_{anchor}, \{\bT_{graph}^{i}\}], ~~~\mathrm{where} \\
    & \bs_{graph}^{i} > 0.5, \ \forall i \in [1,M]. \nonumber
\end{align}

Overall, the anchor proposal module (AnPM) learns to select an important OCR token as an anchor and then construct an ACG for it. 
In this way, AnPM is able to propose a series of ACGs for an input image, which would be fed into the captioning module as guidance to generate diverse captions. 
Meanwhile, the generation process of each ACG is independent and will not be affected by other pairs, which greatly improves the quality of the generated captions.

\subsection{Anchor captioning module}
Compared with general image captioning, the  TextCap requires captioning models to not only describe visual objects but also contain OCR tokens in the generated captions.
To achieve this, we carefully design a progressive captioning module, namely Anchor Captioning Module (AnCM). 
Inspired by the Deliberation Network~\cite{NIPS2017_6775}, as shown in Figure~\ref{fig:overview}, AnCM consists of a visual-captioner (denoted as AnCM$_v$) and a text-captioner (denoted as AnCM$_t$).
First, the visual-captioner, a standard image captioning module, uses the updated visual embedding $\bV$ to generate a visual-specific caption $\mY'$ with $C$ words, denoted as $\mY'=\{ y'_{c}\}_{c=1}^C$.
Then, the text-captioner is proposed to refine the generated caption based on the {text information of ACG} $\mG$ (\textit{see} Eqn.~(\ref{construct_graph})).
Following the training of sequence generation task, the goal of AnCM is to maximise the data log likelihood function as follows: 
\begin{align}
    \mathrm{log} \sum^{C}_{c=1} P(y_c | \mathrm{AnCM}_t(y'_c, \mG)) ~ P(y'_c| \mathrm{AnCM}_v(\bV)), 
\end{align}
where $\{y_c\}$ is the final generated caption.
Since the predicted token $y'_c$ is obtained by argmax function which is a non-differentiable operation, we cannot directly optimise the above equation.
To address this issue, we feed the hidden state feature $\bh_c$ outputted from AnCM$_v$ directly into AnCM$_t$ and the training loss for AnCM$_t$ is computed as follows:
\begin{align}
    \label{eq:AnCM_t}
     \mL_{tcap} = - \mathrm{log} & \sum^{C}_{c=1} P(y_c | \mathrm{AnCM}_t(\bh_c, \mG); \theta_{t}),
     ~~~\mathrm{where} \\
     \label{eq:AnCM_v}
     \bh_c &= \mathrm{AnCM}_v(\bV; \theta_{v}). \nonumber
\end{align}
The $\theta_v$ and $\theta_t$ are learnable parameters of visual-captioner and text-captioner, respectively. {In this way, we can train AnCM in an end-to-end manner}. Next, we will introduce more details about the two captioners.

\noindent\textbf{Visual-captioner ($\mathrm{AnCM}_v$).} 
To capture long-range dependency in sequence modelling, we employ an $L_2$-layer Transformer module ($\Psi$) as the backbone of the visual-captioner. Specifically, the visual-captioner generates tokens in an auto-regressive manner as follows:
\begin{equation}
\begin{split}
    \bh_c &= \Psi(\bV, \mathrm{LM}(\by'_{c-1}); \theta_{v}), \\
     \label{eq:voc_pred}
     y'_c &= \mathrm{argmax} (f_4 (\bh_c)),
\end{split}
\end{equation}
where $\by'_{c-1}$ denotes the embedding of previous output token, $f_4 (\cdot)$ is a linear classifier for common vocabulary and we can obtain the predicted token $y'_c$ with argmax operation. Here, we use the prefix language modelling (LM) technique~\cite{raffel2019exploring} to ensure that the input entries only use previous predictions, and avoid peeping at subsequent generation processes.
Thus far, with the help of the visual-captioner, we obtain a {visual-specific} caption $\{y'_c\}$ and its hidden state feature $\{\bh_c\}_{c=1}^{C}$. Formally, we define the training loss for AnCM$_v$ as $\mL_{vcap}= - \mathrm{log} \sum^{C}_{c=1} P(y'_c)$.

Image captioning is a fairly mature sequence-generation task, and researchers have also proposed many models with promising performance.
In this work, we do not focus on designing a new captioning module. Intuitively, the backbone of visual-captioner can be easily replaced by other image captioning models, such as BUTD~\cite{Anderson2018BottomUpAT} and  AoANet~\cite{Huang2019AttentionOA}.

\noindent\textbf{Text-captioner ($\mathrm{AnCM}_t$).} Based on the hidden state features $\{\bh_c\}_{c=1}^{C}$, the text-captioner aims to generate text-specific captions that contain given OCR tokens. To this end, at this stage, we use ACG as the guidance to refine the caption generated by the visual-captioner from the last step. Specifically, we use an $L_3$-layer Transformer module ($\Psi$) as the backbone of the text-captioner. 
Relying on self-attention, the Transformer allows multimodal embedding to freely interact with others, thereby achieving satisfactory progress in sequence generation tasks. Formally, given the hidden state and ACGs, \wzy{our $\mathrm{AnCM}_t$} can output the joint embedding as follows:
\begin{align}
    \widehat{\mG}, \ \widehat{\by}_c &= \Psi ([\mG, \bh_c, \mathrm{LM} (\by_{c-1})]; \theta_t),
\end{align}
where $\widehat{\mG}$ is the updated token embedding in the ACG and $\widehat{\by}_c$ is the embedding of $c$-th prediction.
Following M4C~\cite{hu2019iterative}, we adopt different classifiers for common vocabulary and OCR candidate tokens as
\begin{align}
    \label{eq:ocr_pred}
    y_c &= \mathrm{argmax}([f_4 (\widehat{\by}_c), \ f_{dp} (\widehat{\mG}, \widehat{\by}_c)]),
\end{align}
where $f_4$ is the shared classifier with visual-captioner in Eqn.~(\ref{eq:voc_pred}), $f_{dp}$ denotes the dynamic pointer network \cite{hu2019iterative} that makes prediction based on the $\widehat{\mG}$ and $\widehat{\by}_c$.
After concatenating two prediction scores, we use the argmax function on the final prediction score to obtain the predicted token $y_c$.

Compared with general captioning methods, the proposed AnCM makes full use of a key character of the TextCap task, that is, the OCR token can be used as an important guide to improve the generation accuracy. 

\subsection{Training details}
\mata{Formally, we train our \fullname{} model by optimising the following loss function:
\begin{align}
    \label{eq:over_loss}
    \mL =  & \ \mL_{anchor} (\bs_{anchor})+ \alpha \ \mL_{graph} (\bs_{graph}) \\
    & + \ \beta \ \mL_{vcap} (\mY') + \eta \ \mL_{tcap} (\mY), \nonumber
\end{align}
where {the $\bs_{anchor/graph}$ is the output of AnPM (\textit{see} Eqn.~(\ref{eq:anchor_select}) or (\ref{eq:graph_cons})), the $\mY'\small{=}\{y'_{c}\}$ and $\mY\small{=}\{y_{c}\}$ denote the generated visual-specific caption and text-specific caption (\textit{see} Eqns.~(\ref{eq:voc_pred}) and (\ref{eq:ocr_pred})), respectively.}}
$\{\alpha,\beta,\gamma\}$ are trade-off parameters.
In practice, all the above four losses adopt the binary cross-entropy loss function.
We train AnPM with $\mL_{anchor}$ and $\mL_{graph}$ to find the most frequently described ACGs.
We train AnCM with $\mL_{vcap}$ and $\mL_{tcap}$ to generate visual-specific and text-specific captions. Due to the page limitation, we put detailed training and inference algorithms in Supplementary A.

\noindent\textbf{Ground-truth labels.} 
Next, we will illustrate how to obtain supervisions for training AnPM and AnCM. 
1) Given a manually annotated caption (\eg `\textit{a man ... number \underline{15} on it}' in Figure~\ref{fig:overview}), we consider it as ground truth (gt) for $\mL_{tcap}$ to train the text-captioner. 
2) We then mask the OCR tokens contained in the manually generated caption with \underline{[unk]} to build a new caption (\eg `\textit{a man ... number \underline{[unk]} on it}') for $\mL_{vcap}$ to train the visual-captioner. This is because we do not require the AnCM$_v$ to make prediction among OCR tokens by using only the visual information. 
3) Considering that different people will describe the same image from different views, the most frequently described token is often the most important one for describing an image. Thus, we choose the most frequently described token as gt for $\mL_{anchor}$ to train the anchor prediction module in AnPM.
4) Given the chosen anchor, we consider the tokens that appear in the same caption as gt for $\mL_{graph}$ to train AnPM to find the most relevant tokens for an anchor.
During the test phase, we do not need to construct gt-ACGs since they are only used for calculating losses in the training. Note that the gt-ACGs are automatically mined and constructed from the same training split without using any additional annotations, and thus comparisons with other methods are fair.

\section{Experiments}
We verify the effectiveness of our method on the TextCaps~\cite{Sidorov2020TextCapsAD} dataset. In the following, we first briefly introduce the TextCaps and the comparison settings for it in Sec.~\ref{dataset}. 
More implementation details can be found in the Sec.~\ref{implementation}.
And then, we compare our method with existing captioning models in Sec.~\ref{main_exp} and Sec.~\ref{abl_exp}. Last, we demonstrate our proposed method by providing some visualisation results and analysis in Sec.~\ref{sec:visual}.
\subsection{Datasets and settings}
\label{dataset}
\noindent\textbf{Datasets.} The TextCaps dataset~\cite{Sidorov2020TextCapsAD} is collected from Open Image V3 dataset and contains 142,040 captions on 28,408 images, which have been verified to contain text through the Rosetta OCR system~\cite{borisyuk2018rosetta} and human annotators.
For each image, there are five independent captions. In the test split, each image has an additional caption \wzy{that} is collected to estimate human performance on the dataset. 
The dataset also contains captions where OCR tokens are not presented directly but are used to infer a description~\cite{Sidorov2020TextCapsAD}. 
In this case, the captioning models are required to perform challenging reasoning rather than simply copy the OCR tokens. 
Most captions contain two or more OCR tokens, and the average length of captions is 12.4.
\begin{table}[t]
\small
\resizebox{0.48\textwidth}{!}{
\begin{tabular}{clccccc}
\hline
\multirow{2}{*}{\#}     & \multicolumn{1}{c}{\multirow{2}{*}{Method}} & \multicolumn{5}{c}{TextCaps validation set metrics}                                                                                                                                        \\ \cline{3-7} 
                        & \multicolumn{1}{c}{}                        & B                             & M                             & R                            & S                               & C                               \\ \hline
1                       & BUTD                                & 20.1                                & 17.8                                & 42.9                                & 11.7                                & 41.9                                \\
2                       & AoANet                                & 20.4                                & 18.9                                & 42.9                                & 13.2                                & 42.7                                \\
3                       & M4C-Captioner                                & 23.3                                & 22.0                                & 46.2                                & 15.6                                & 89.6                                \\
4                       & M4C-Captioner$^{-}$                                        & 15.9                                & 18                                & 39.6                                & 12.1                                & 35.1                                \\
5                       & AnCM$_v$                                & 16.1                                & 16.3                                & 40.1                                & 11.2                                & 29.1                                \\
6                       & \textbf{Ours}                          & \textbf{24.7}                      & \textbf{22.5}                      & \textbf{47.1}                      & \textbf{15.9}                      & \textbf{95.5}                      \\ \hline
\multirow{2}{*}{\#}       & \multicolumn{1}{c}{\multirow{2}{*}{Method}} & \multicolumn{5}{c}{TextCaps test set metrics}                                                                                                                                        \\ \cline{3-7} 
                        & \multicolumn{1}{c}{}                        & B                             & M                             & R                            & S                               & C                               \\ \hline
7                       & BUTD                                & 14.9                                & 15.2                                & 39.9                                & 8.8                                & 33.8                                \\
8                       & AoANet                                & 15.9                                & 16.6                                & 40.4                                & 10.5                                & 34.6                                \\
\multicolumn{1}{c}{9} & M4C-Captioner                                & \multicolumn{1}{c}{18.9}           & \multicolumn{1}{c}{19.8}           & \multicolumn{1}{c}{43.2}           & \multicolumn{1}{c}{12.8}           & \multicolumn{1}{c}{81.0}           \\
\multicolumn{1}{c}{10} & MMA-SR                         & \multicolumn{1}{c}{19.8} & \multicolumn{1}{c}{20.6} & \multicolumn{1}{c}{44.0} & \multicolumn{1}{c}{13.2} & \multicolumn{1}{c}{\textbf{88.0}} \\
\multicolumn{1}{c}{11} & \textbf{Ours}                          & \multicolumn{1}{c}{\textbf{20.7}} & \multicolumn{1}{c}{\textbf{20.7}} & \multicolumn{1}{c}{\textbf{44.6}} & \multicolumn{1}{c}{\textbf{13.4}} & \multicolumn{1}{c}{87.4} \\ \hline
\multicolumn{1}{c}{12} & Human                                & \multicolumn{1}{c}{24.4}           & \multicolumn{1}{c}{26.1}           & \multicolumn{1}{c}{47.0}           & \multicolumn{1}{c}{18.8}           & \multicolumn{1}{c}{125.5}           \\\hline
\end{tabular}}

\caption{Comparison with SOTA methods on the validation and test set. In particular, rows 4 are captioning models without OCRs, ~\ie~only use visual information to generate captions. The last row is the estimated human performance, which can be seen as the upper bound of captioning models on the TextCaps dataset.}
\label{tab:main_results}
\end{table}

\noindent\textbf{Evaluation metrics.} We use five standard evaluation metrics in image captioning, including BLEU (B)~\cite{BLEU}, METEOR (M)~\cite{Meteor}, ROUGE\_L (R)~\cite{ROUGE}, SPICE (S)~\cite{SPICE} and CIDEr (C)~\cite{CIDEr} to evaluate the accuracy. Following the benchmark setting~\cite{Sidorov2020TextCapsAD}, we mainly focus on CIDEr, which puts more weight on informative tokens and is more suitable for this dataset. To evaluate the diversity of generated captions, we use Div-n~\cite{Li2016ADO} and SelfCIDEr~\cite{Wang2019DescribingLH} metrics on the validation set. In particular, the Div-n focuses on token-level diversity while the SelfCIDEr is used for semantic-level diversity. 
In addition, we propose a new metric, called Cover Ratio (CR), to measure the content diversity, that is, how many OCR tokens are included in the generated captions. 
{For notation convenience, we {omit} the percentage in the metric scores.}

\noindent\textbf{Compared methods.} We first compare our method with two state-of-the-art (SOTA) image captioning methods, \ie~BUTD~\cite{Anderson2018BottomUpAT} and AoANet~\cite{Huang2019AttentionOA}. For the TextCap task, we compare our method with the current SOTA methods M4C-Captioner~\cite{Sidorov2020TextCapsAD} \rev{and MMR-SA~\cite{MMA2020Wang}}. For fair comparisons, {we use the same dataset annotation and multimodal feature extraction methods (including the OCR system and Faster RCNN) for all considered methods in our experiments.}
We conduct extensive experiments on the validation and test set. In particular, the evaluation results are provided by the test server of the TextCaps-Challenge\footnote{TextCaps: \href{https://textvqa.org/textcaps}{https://textvqa.org/textcaps}}. Since the number of submissions of the results on the test set is limited, we conduct ablation studies on the validation set.

\subsection{Implementation details}
\label{implementation}

In our implementation\footnote{\href{https://github.com/guanghuixu/AnchorCaptioner}{https://github.com/guanghuixu/AnchorCaptioner}.}, the feature dimension $d$ is 768 and the $f_{*}$ is a linear layer with LayerNorm activation function to stabilise training. We train our model for 12,000 iterations with a batch size of 128. During training, we use the Adamax optimiser~\cite{kingma2014adam} with a learning rate of 2e-4. 
We adopt default parameter settings of BERT-BASE~\cite{devlin2018bert} for the transformer module $\Psi$, such as 12 self-attention heads. But the number of stacked layers are $L_1=2, L_2=L_3=4$, respectively. For fair comparisons, following the TextCaps benchmark~\cite{Sidorov2020TextCapsAD}, we use the same fixed common vocabulary and the feature embeddings of visual objects and OCR tokens. The number of visual objects is $N=100$ and the number of OCR tokens is $M=50$. The maximum generation length is $C=30$.
The trade-off parameters of different losses are set to $\alpha=\beta=\gamma=1$.

\subsection{Main results}\label{main_exp}

\begin{table}[t]
\small

\begin{tabular}{clcccc}
\hline
\#                      & Method & Div-1                 & Div-2                 & selfCIDEr     & CR   \\ \hline
1                       & BUTD                        & 27.0                     & 36.4                     & 45.3       & -            \\
2                       & M4C-Captioner               & 27.2                     & 41.2                     & 49.4          & 27.3        \\
3  & \textbf{Ours}            & \textbf{29.8} & \textbf{43.8} & \textbf{57.6}   & \textbf{37.8} \\ \hline
4  & Human     & 62.1 & 87.0 & 90.9 & 19.3
\\ \hline
\end{tabular}

\caption{Diversity analysis. The BUTD and M4C-Captioner generate diverse captions via beam search  (beam size is 5).}
\label{tab:diversity}
\end{table}
\begin{table}[t]
\small

\resizebox{0.48\textwidth}{!}{
\begin{tabular}{clccccccc}
\hline
\# & Projection & B    & M    & R   & S      & C      & A & F1 \\ \hline
1                   & Single            & 23.9          & 22.2          & 46.7          &  15.6          &  90.3    & 48.4 & 68.8    \\
2                   & Multiple            & 23.7          & 22.4          & 46.3          &  16.0          &  90.7    & 49.0 & 68.9    \\
3                   & \textbf{Sequence}                                & \textbf{24.7 } & \textbf{22.5} & \textbf{47.1} & \textbf{15.9} & \textbf{95.5} &  \textbf{49.1}    & \textbf{71.8}   \\ \hline
\end{tabular}}

\caption{Ablation studies of anchor proposal module (AnPM) with independent projection (FC) and sequence projection (RNN). Apart from using captioning metrics, we also use accuracy (A) and F1 score to further measure the performance of AnPM.}
\label{tab:apm}
\end{table}

\noindent\textbf{Overall results.} As shown in Table~\ref{tab:main_results}, we first compare our method with the current SOTA captioning models, including BUTD, AoANet, and M4C-Captioner. 
From the table, the BUTD and AoANet, standard image captioning models, show poor performances on the validation set since they fail to describe the texts in images. 
M4C-Captioner reasons over multimodal information and outperforms standard image captioning models by a large margin. 
Compared with M4C-Captioner, our model improves the CIDEr score from 89.6 to 95.5 on the validation set and achieves 6 absolute improvement on the test set.
In particular, we also report the result of visual-captioner AnCM$_v$ (row 5), which can be seen as a degraded version of our model without using OCR tokens. 
Same as M4C-Captioner w/o OCRs (row 4), AnCM$_v$ is hard to generate reliable captions for the TextCaps dataset. 
To address this issue, our model is equipped with an additional text-captioner that refines generated captions with the text information. {For fair comparisons, we choose the ACG with the highest anchor score to refine the generated caption in this experiment, since existing methods derive only one caption for each input image.} In this way, our full model further boosts the CIDEr score from 29.1 to 95.5 in the validation set.

\noindent\textbf{Diversity analysis.} 
{To further evaluate the diversity of the generated captions,}
we compare \fullname{} with BUTD and M4C-Captioner in terms of diversity metrics. Since existing methods only generate a global description for an image, we use the beam search technique for them to produce diverse captions as baselines, where the beam size is set to 5. 
{For fair comparisons, in our method, we also sample five ACGs for each image to generate captions.}
As shown in Table~\ref{tab:diversity}, our method surpasses baselines in terms of all  considered metrics. Interestingly, the ground-truth captions (by humans) have high selfCIDEr but with low OCR cover ratio (CR). It means that humans may tend to describe the salient image contents but ignore some OCR tokens. Compared with human captioning, our method is able to {generate multiple captions with content diversity, covering more OCR tokens.} Note that, cover ratio (CR) score for BUTD method is empty, because OCR tools are not used in it.

\subsection{Ablation studies}
\label{abl_exp}
In this section, we further conduct ablation studies to demonstrate the effectiveness of AnPM and AnCM.

\begin{table}[t]
\small
\resizebox{0.48\textwidth}{!}{
\begin{tabular}{cccccccc}
\hline
\# & Anchor                                        & ACG & B & M & R & S & C  \\ \hline
1  & \multirow{3}{*}{Large}                        & All      & 21.2   & 21.0   & 44.8    & 14.5 & 76.6  \\  
2  &                                               & Around & 21.4   & 21.1   & 44.9    & 14.4  & 77.4  \\  
3  &                                               & Random   & 20.8   & 20.7    & 44.4    & 14.1  & 72.6   \\ \hline
4  & \multirow{3}{*}{Centre}                       & All      & 21.2   & 21.0    & 44.8    & 14.5 & 76.6  \\  
5  &                                               & Around & 21.5    & 21.2   & 45.0    & 14.4  & 78.0  \\  
6  &                                               & Random   & 20.7    & 20.8   & 44.5    & 14.1 & 73.1  \\ \hline
7  & \multirow{2}{*}{-} & All      & 21.1    & 21.1    & 44.7     & 14.6 & 76.7  \\  
8  &                                               & Random   & 20.4    & 20.6   & 44.1    & 13.9  & 70.2  \\ \hline
9  & \multirow{3}{*}{GT}                           & All      & 23.5   & 22.4   & 46.3    & 15.7  & 90.3  \\  
10 &                                               & Around & 22.1   & 21.9   & 45.6    & 15.2 & 83.9   \\  
11 &                                               & Random   & 21.4   & 21.2   & 45.0       & 14.7 & 78.7  \\ \hline
12 & AnPM                                      & AnPM & 24.7    & 22.5   & 47.1     & 15.9 & 95.5  \\
13 & \textbf{GT}                                            & \textbf{GT}       & \textbf{25.6}   & \textbf{23.4 }   & \textbf{48.1}    & \textbf{16.9 } & \textbf{104.9} \\ \hline
\end{tabular}}
\caption{Ablation studies of ACG construction using rule-based approaches. For instance, row 2 ('Large'+'Around') means that we choose an OCR token with the largest region size as the anchor, and then group the five closest tokens to construct its ACG. In particular, we randomly group some tokens into an ACG, denoted as `Random' in the table. `GT' denotes ground-truth and 'AnPM' means using the prediction of AnPM.}
\label{tab:rule}
\end{table}
\begin{table}[t]
\small

\begin{tabular}{clccccc}
\hline
\# & Method & B    & M    & R   & S      & C      \\ \hline
1                       & M4C-Captioner                                & 23.3                                & 22.0                                & 46.2                                & 15.6                                & 89.6                                \\
2                   & M4C-Captioner$^{\dagger}$                               & 24.1      & 22.6      & 46.7      & 15.7      & 93.8      \\
3                   & M4C-Captioner$^{*}$                               & 24.4      & 22.6      & 46.9      & 15.8      & 99.6      \\ \hline
4                   & AnCM$_v$ + AnCM$_t^{\dagger}$                                & {24.7} & {22.5} & {47.1} & {15.9} & {95.5} \\
5                   & AnCM$_v$ + AnCM$_t^{*}$                               & \textbf{25.6} & \textbf{23.4} & \textbf{48.1} & \textbf{16.9} & \textbf{104.9} \\ \hline
\end{tabular}

\caption{Ablation studies of Anchor Caption Module (AnCM). $\dagger$ denotes captioning modules using prediction ACGs provided by AnPM, while $*$ denotes captioning modules using ground-truth.}
\label{tab:acm}
\end{table}
\begin{figure*}[h]
    \centering
    \includegraphics[width=\textwidth]{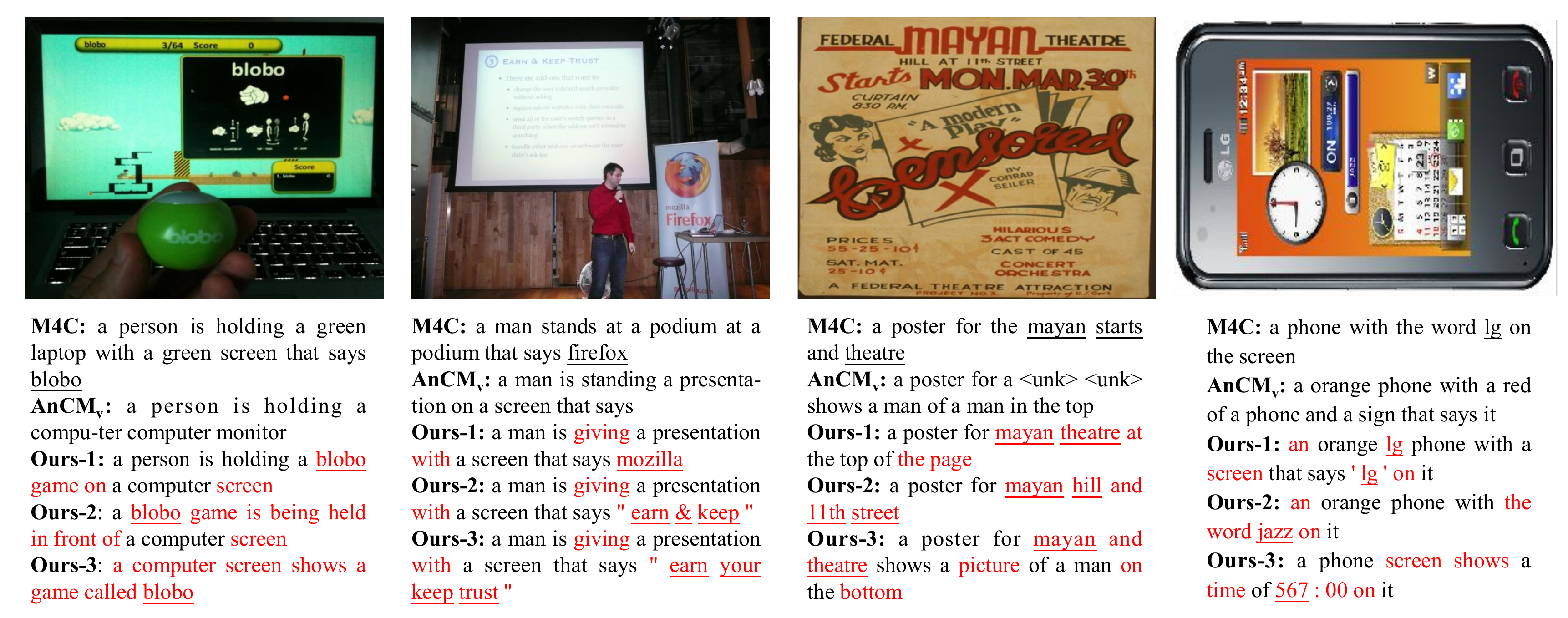}
    \caption{Visualisation results on the TextCaps validation set. The prediction results of M4C-Captioner (M4C), visual-captioner (AnCM$_v$) and the proposed \fullname{} are placed below the images in turn. The $<$unk$>$ denotes `unkown' token. For better visualisation, the underlined word is copy from OCR tokens. In particular, \fullname{} will refine the caption generated by AnCM$_v$. The modified tokens are viewed in \textcolor{red}{red} colour.}
    \label{fig:visualization}
\end{figure*}

For AnPM, we first compare \rev{three} different kinds of ACG construction strategies,~\ie~independent projection (FC), multiple projection (transformer module) and sequence projection (RNN module). As shown in Table~\ref{tab:apm}, the \textbf{RNN} outperforms \textbf{FC} in terms of all considered metrics, especially improves the CIDEr score from 90.3 to 95.5. As discussed in Sec.~\ref{sec:apm}, the sequence projection is more reasonable since it considers the history prediction. \rev{More details can be found in the supplement material.} 
Moreover, we also report the accuracy of anchor prediction and the F1 score of the predicted ACG.
Note that, there is a trade-off between obtaining high F1 score and diversity. To achieve high accuracy and F1 score, AnPM tends to predict the most frequently described ACG, which, however, could suffer from low diversity of generated captions. 

In addition to the above comparisons, we also compare AnPM (RNN projection) with the rule-based ACG construction and report the quantitative results in Table~\ref{tab:rule}. 
To be specific, we first adopt different rules to select token as an anchor, including the largest token (rows 1-3), the centre token (rows 4-6),  the ground-truth anchor (rows 9-11). Then, we choose tokens to construct ACG using different strategies (\ie~`All / Around / Random'). In particular, we try to group tokens into a graph directly without performing anchor selection process (in rows 7-8).
From the table, all the rule-based methods suffer low metric performance even given the GT anchor to construct ACGs. The learning-based method (AnPM) outperforms rule-based methods by a large margin. One reason is that our AnPM considers the rich semantic information of the tokens themselves and the visual information in images, while the rule-based approaches mainly use shallow information such as size and location.

We also conduct ablation experiments for AnCM. From the results in Table~\ref{tab:acm}, we draw the following main observations. 1) As shown in rows 1-3, the M4C-Captioner$^{\dagger}$ and M4C-Captioner$^*$ that take the predicted ACGs and ground-truth ACGs as inputs, outperform the original M4C-Captioner by around 4 and 10 in terms of the CIDEr score, respectively. These results well verify our idea, \ie~first group OCR tokens into \wzy{different ACG} and then describe each ACG with a specific caption.
2) Compared with M4C-Captioner (row 1), our method improves CIDEr score from 89.6 to 95.5.
3) Equipped with AnPM, the M4C-Captioner$^{\dagger}$ (row 2) achieves better performance, which implies that our AnPM can be easily extended to existing text-based reasoning methods.
4) Even for the same ACG inputs, our method is still superior to M4C-Captioner$^{\dagger}$ and M4C-Captioner$^{*}$, which demonstrates the powerful captioning ability of our AnCM.
\wzy{5) According to the last two rows, our AnCM suffers a performance degradation with the predicted ACGs as input, indicating that our method still has great potentials for improving.}

\subsection{Visualisation analysis}
\label{sec:visual}
To further demonstrate the effectiveness of our method, we show some visualisation results on the TextCaps validation set. From Figure~\ref{fig:visualization}, our \fullname{} is able to refine the rough captions generated by the visual-captioner (AnCM$_v$). Specifically, for each input image, AnCM$_v$ first uses visual information to generate a global caption, such as `a man' and `a poster'. Similar to general image captioning models, AnCM$_v$ \wzy{is difficult to describe the texts in images.}
As a result, we can see the {visual-specific} captions may contain some $<$unk$>$ tokens. It means that AnCM$_v$ cannot use limited information to generate reasonable predictions in this case. And then, \fullname{} use anchor-centred graphs (ACGs) to further refine the {visual-specific} captions. Note that, the refine process is not only to replace the $<$unk$>$ token, but also to revise the entire caption. 
\rev{There are 66.39\% of generated captions with $<$unk$>$, and each caption has 1.24 $<$unk$>$ on average. AnCM$_t$ modified 26.85\% of words on the AnCM$_v$'s output and improved CIDEr score from 29.1 to 95.5 (\textit{see} Tabel~\ref{tab:main_results}).}
We also randomly sample different ACGs to demonstrate the diversity of our generation. Compared with M4C-Captioner, our method is able to generate fine-grained captions and cover more OCR tokens. \rev{To further demonstrate the controllability and diversity of our method, we provide more visualisation results in the supplement material.}

\section{Conclusion}
In this paper, we have proposed an \fullname{} to resolve the TextCap task.  \rev{To solve this task, existing methods tend to generate only one rough global caption which contains one or two salient objects in the complex scene. Intuitively, such methods may ignore some regions that people are really interested in. Unlike existing methods}, we seek to generate multiple captions from different views \rev{and cover more valuable scene information. Specifically}, we first propose an anchor proposal module to 
group OCR tokens and construct anchor-centred graphs (ACGs) by modelling the relationship between image contents. After that, our anchor captioning module first generates a rough visual-specific caption and then uses the above ACGs to further refine it to multiple text-specific captions. In this way, our method is able to generate diverse captions to cover more information in images. Our method achieves state-of-the-art performance on the TextCaps dataset and outperforms the benchmark by 6 in terms of CIDEr score. Extensive ablation experiments also verify the effectiveness of each component of our method. 
Note that our anchor captioning module has the potential to solve both image captioning and text-based image captioning tasks simultaneously, which we leave to our future study.

\noindent\textbf{Acknowledgements.}
This work was partially supported by the Science and Technology Program of Guangzhou, China, under Grant 202007030007, Fundamental Research Funds for the Central Universities D2191240, Program for Guangdong Introducing Innovative and Enterpreneurial Teams 2017ZT07X183, Opening Project of Guangdong Key Laboratory of Big Data Analysis and Processing.

{\small
\bibliographystyle{ieee_fullname}
\bibliography{egbib}
}



\def\fullname{Anchor-Captioner}
\def\shortname{A-Cap}
\renewcommand {\thetable} {S\arabic{table}}
\renewcommand {\thefigure} {S\arabic{figure}}
\def\mata{\textcolor{magenta}}
\def\rev{\textcolor{black}}



\def\cvprPaperID{1215} 
\def\confYear{CVPR 2021}


\title{Supplementary Materials: Towards Accurate Text-based Image Captioning \\ with Content Diversity Exploration}

\author{
    Guanghui Xu$^{1,2}$\thanks{Authors contributed equally.},
    Shuaicheng Niu$^{1*}$, Mingkui Tan$^{1,4}$, Yucheng Luo$^{1}$, Qing Du$^{1,4}$\thanks{Corresponding author}, Qi Wu$^{3}$ \\
    $^{1}$South China University of Technology,
    $^{2}$Pazhou Laboratory, $^{3}$ University of Adelaide \\
    $^{4}$Key Laboratory of Big Data and Intelligent Robot, Ministry of Education, \\
    {\tt\small{sexuguanghui}@mail.scut.edu.cn},
    {\tt\small \{mingkuitan, duqing\}@scut.edu.cn}, {\tt\small{qi.wu01@adelaide.edu.au}}
}
\onecolumn
\maketitle
\appendix

\begin{center}
	{
		\Large{\textbf{Supplementary Materials}}
	}
\end{center}

In the supplementary, we provide more implementation details and experimental results of the proposed \fullname{}. We organise the supplementary as follows.
\vspace{-3pt}
\begin{itemize}
    \item In Section~\ref{algorithms}, we provide the detailed training and inference algorithms of  \fullname{}.
    \vspace{-3pt}
    \item In Section~\ref{detailed_acg}, we give more discussions on the anchor-centred graph (ACG) construction strategy.
    \vspace{-3pt}
    \item In Section~\ref{ablation}, we conduct more ablation experiments to verify the generalisation ability of \fullname{}.
    \vspace{-3pt}
    \item \rev{In Section~\ref{sec:loss}, we conduct more ablation experiments to further measure the importance of each loss term.}
    \vspace{-3pt}
    \item In Section~\ref{visualisation}, we show more visualisation results to further verify the promise of the proposed method. 
\end{itemize}

\section{Algorithms}\label{algorithms}
\vspace{-15pt}
\begin{figure*}[htp]
	\begin{minipage}[t]{0.48\linewidth}
		\centering
		\begin{algorithm}[H]
		\small
		\caption{Training Method of \fullname}
		\label{training_algorithm}
		\begin{algorithmic}[1]
			\REQUIRE Multimodal feature set $\{\widehat{\bV},\widehat{\bT}\}$, transformer encoder $\Psi$, anchor predictor $\phi$, sequence projection RNN, visual captioner $\text{AnCM}_v$, text captioner AnCM$_t$, overall model parameters $\theta$.\\
			\STATE Initialize the model parameters $\theta$.\\
			\WHILE{not converge}
			\STATE Randomly sample a feature pair $(\widehat{\bV},\widehat{\bT})$.
			\STATE ${\bV}, {\bT} = \Psi(\widehat{\bV}, \widehat{\bT}; \theta_{a})$. // \emph{Multimodal Embedding Fusion}
			\STATE // \emph{Anchor Proposal Module} \\
			$\bs_{anchor} = \mathrm{Softmax} \ (\phi({\bT}))$ ~~~// \emph{Predict anchor scores}. \\
			Choose the token with the highest score as $\bT_{anchor}$. \\
			Construct anchor-centred graph $\mG$ for $\bT_{anchor}$ using RNN. \\
			\STATE // \emph{Anchor Captioning Module} \\
			$\bh_c = \mathrm{AnCM}_v({\bV}, \by'_{c-1}; \theta_{v})$ ~~~// \emph{Obtain global visual info}. \\
			$\widehat{\mG}, \widehat{\by}_c = \mathrm{AnCM}_t (\mG, \bh_c, \mathrm{LM} (\by_{c-1}); \theta_t)$ ~~~// \emph{refine.} \\
			Obtain the rough caption $\mY'=\{y'_c\}$ using Eqn.~(8). \\
			Obtain the fine-grained caption $\mY=\{y_c\}$ using Eqn.~(10).\\
			\STATE Update $\theta$ using overall training loss (Eqn.~11)
			\ENDWHILE
		\end{algorithmic}
	\end{algorithm}
	\end{minipage}\hfill
	\begin{minipage}[t]{0.48\linewidth}
	\centering
	\begin{algorithm}[H]
		\small	
		\caption{Inference of \fullname}	\label{inference_algorithm}
		\begin{algorithmic}[1]
			\REQUIRE multimodal features ($\widehat{\bV},\widehat{\bT}$), transformer encoder $\Psi$, anchor predictor $\phi$, sequence projection RNN, visual captioner AnCM$_v$, text captioner AnCM$_t$; the number of sampled ACGs ($K$), pretrained model parameters $\theta$.\\
			\STATE // \emph{Multimodal Embedding Fusion} \\
			${\bV}, {\bT} = \Psi(\widehat{\bV},\widehat{\bT}; \theta_{a})$.
			\STATE // \emph{Anchor Proposal Module} \\
			$\bs_{anchor} = \mathrm{Softmax} \ (\phi({\bT}))$ ~~~// \emph{Predict anchor scores}. \\
			Choose top-$K$ tokens as the anchors $\{\bT_{anchor}^{k}\}_{k=1}^{K}$. \\
			Construct ACGs $\{\mG_k\}$ for the anchors using RNN.
			\STATE // \emph{Anchor Captioning Module} \\
			$\bh_c = \mathrm{AnCM}_v({\bV}, \by'_{c-1}; \theta_{v})$ ~~~// \emph{Obtain global visual info}.\\
			\FOR{$k$ = $1,\dots,K$} 
			\STATE $\widehat{\mG}_k, \widehat{\by}_c^{k} = \mathrm{AnCM}_t (\mG_k, \bh_c, \mathrm{LM} (\by_{c-1}^{k}); \theta_t)$ \\
			Obtain the rough caption $\mY'_k=\{y'_c\}_k$ using Eqn.~(8). \\
			Obtain the $k$-th caption $\mY_k=\{y_c\}_k$ using Eqn.~(10).
			\ENDFOR
			\STATE Obtain $K$ rough captions and $K$ fine-grained captions
		\end{algorithmic}
	\end{algorithm}
	\end{minipage}
\end{figure*}
In this section, we provide the detailed training and inference algorithms of our method in Algorithms~\ref{training_algorithm} and~\ref{inference_algorithm}. Given an input image, we first fuse visual and text features to obtain multimodal embeddings. Then, we apply the anchor proposal module (AnPM) to choose and group texts to construct a series of anchor-centred graphs (ACGs), where each ACG denotes a group of relevant OCR tokens that are used to generate a specific caption.
Last, we employ the visual-
captioner (AnCM$_v$) to generate a rough caption and then use ACGs as guidance to refine the generated caption by the text-captioner (AnCM$_t$). In particular, we adopt the top-1 ACG for the training while using top-K ACGs to generate K diverse captions in the inference.

\section{More details about ACG}\label{detailed_acg}
\begin{figure*}[t]
    \renewcommand\thefigure{S1}
    \centering
    \includegraphics[width=\linewidth]{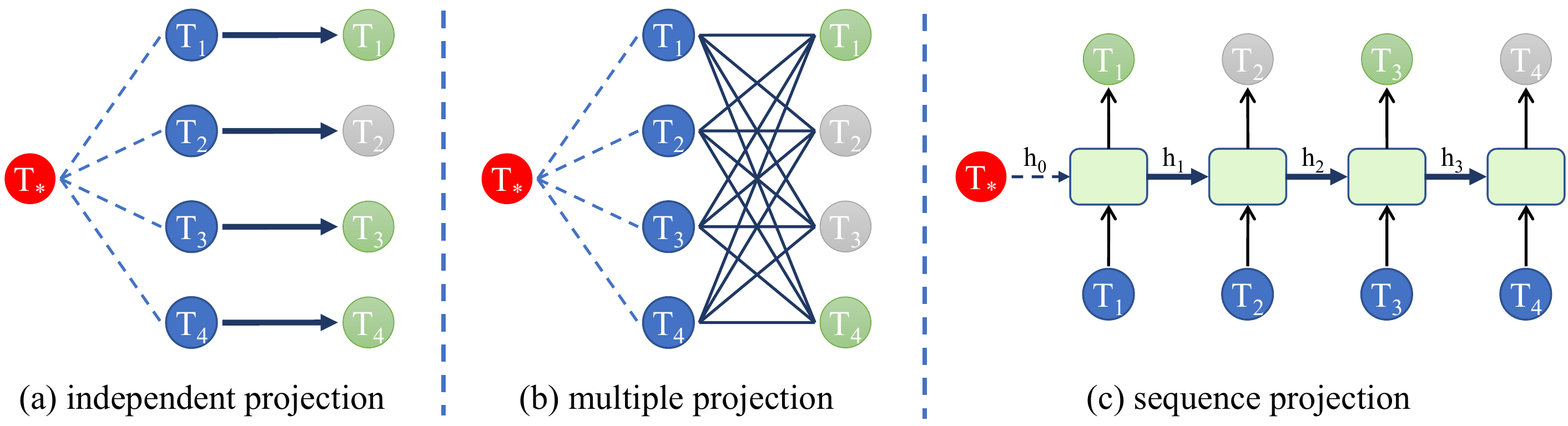}
    \caption{Illustrations of different ACG construction strategies. The red circle denotes the anchor, and the blue circles denote the candidate OCR tokens to be selected. After performing prediction, the green one indicates that the current token and the anchor belong to the same group of ACG, while the gray is the opposite. \rev{It means that the ACGs in (a-c) are $\{\mathrm{T_*, T_1, T_3, T_4}\}$, $\{\mathrm{T_*, T_1, T_4}\}$ and $\{\mathrm{T_*, T_1, T_3}\}$, respectively.}
    }
    \label{fig:projection}
\end{figure*}
As shown in Figure~\ref{fig:projection}, to construct an anchor-centred graph (ACG), 
\rev{we mainly consider three} kinds of construction strategies,~\ie~independent projection (fully connected layer), \rev{multiple projection (4-layer Transformer module)} and sequence projection (RNN module). 
Specifically, based on a given anchor $\bT_{anchor}$,
the independent projection directly predicts a correlation score for each token without considering others, \rev{while the multiple projection considers global information via self-attention mechanism. As shown in Figure~\ref{fig:projection}, the multiple projection makes prediction for a token (\eg{$\bT_1$}) using rich neighbourhood information. In particular, the sequence projection considers the relationships among $\bT_{anchor}$, all OCR tokens and history predictions. The order of OCR tokens is determined by the ranking of confidence scores (in descending order) obtained from the OCR model. In a word, the three strategies use different information to construct ACGs, where the independent and multiple projection mainly consider local and global information, and the sequence projection perceives previous predictions through hidden state.}
To some extent, the sequence projection will be more reasonable, because the choice between different tokens to construct an ACG is not completely independent.
To train AnPM, we first parse ground-truth (gt) captions into candidate sets, where gt-anchor is the most frequently described token and gt-ACG are the tokens appeared in the same caption. 
Here, we do not have semantic labels of the visual objects in an image, and thus we have trouble to know what objects are included in the ground-truth captions. In this work, we only consider the OCR token and its relative tokens to construct ACGs. We will consider training AnPM to propose sub-regions as RPN in the future study.

\section{More ablation studies about generalisation}\label{ablation}
\begin{table*}[t]
\renewcommand\thetable{S1}
\centering
\begin{tabular}{clccccccc}
\hline
\# & Method                         & trained on                     & evaluated on & BLEU  & METEOR & ROUGE\_L & SPICE & CIDEr \\ \hline
1  & \multirow{6}{*}{M4C} & \multirow{2}{*}{COCO}          & COCO         & 34.3  & 27.5   & 56.2     & 20.6  & 112.2 \\  
2  &                                &                                & TextCaps     & 12.3  & 14.2   & 34.8     & 9.2   & 30.3  \\ \cline{3-9} 
3  &   & \multirow{2}{*}{TextCaps}                       & COCO     & 8.3  &  15.1 &   34.2   &  8.0 & 17.3  \\
4  &   &          & TextCaps     & 23.3  & 22.0   & 46.2     & 15.6  & 89.6  \\ \cline{3-9} 
5  &                                & \multirow{2}{*}{COCO+TextCaps} & COCO         & 27.1  & 24.1   & 51.6     & 17.4  & 87.5     \\  
6 &     &   & TextCaps     & 21.9  & 22.0     & 45.0  & 15.6  & 84.6 \\ \hline
7  & \multirow{6}{*}{Ours}          & \multirow{2}{*}{COCO}          & COCO         &  34.6     &   27.3     &  56.1        &   20.2    &   110.3    \\ 

8  &      &                                & TextCaps     &   12.6    &  13.8      &    35.2      &   8.8    & 29.2       \\ \cline{3-9} 
9  &                                & \multirow{2}{*}{TextCaps}  
& COCO         &   8.9    &   15.5     &   34.7       &  8.3     & 18.4      \\  
10  &                                &                                & TextCaps     & 24.7  & 22.5   & 47.1     & 15.9  & 95.5  \\ \cline{3-9} 
11  &                                & \multirow{2}{*}{COCO+TextCaps} & COCO         & 30.5  &  25.2      &   53.6       &  18.4     &  96.3     \\  
12 &                                &                                & TextCaps     & 23.6 & 22.2  & 46.2     & 15.7  & 90.0  \\ \hline
\end{tabular}
\caption{More experiments about generalisation. We train our model and M4C-Captioner on TextCaps and COCO captioning training split and then evaluate the models on the different validation split. Specifically, `COCO+TextCaps' denotes that a model uses both COCO captioning and TextCaps dataset for joint training. In practice, since the scale of COCO is much larger than TextCaps, we set the sampling rate to 1:8 to sample TextCaps as frequently as COCO.}
\label{tab:s1}
\end{table*}

To further demonstrate the generalisation ability of our method, we conduct ablation experiments on COCO captioning and TextCaps dataset. COCO captioning is a famous large-scale dataset for general image captioning and TextCaps dataset is recently proposed to enhance the captioning ability of existing methods, especially the reading ability.
As discussed in the main paper, general image captioning methods mainly focus on visual objects and overall scenes in images, while ignoring text information that is of critical importance for comprehensively understanding images. In this sense, it is necessary to study generalisation ability of existing methods on COCO captioning and TextCaps dataset. To this end, we exploit different settings to conduct experiments.
From the results in Table~\ref{tab:s1}, we draw the following main observations: 1) When only training models on COCO captioning (rows 1-2 and 7-8), our model achieves comparable performance as M4C-Captioner.
2) When training models on TextCaps dataset (rows 3-4 and 9-10), our model outperforms M4C-Captioner in terms of two evaluation settings. 3) When jointly training models using both COCO and TextCaps dataset (rows 5-6 and 11-12), our model improves the CIDEr score from 87.5\% to 96.3\% on COCO and achieves 5\% absolute improvement on TextCaps. 4) Unfortunately, as shown in rows 5-6 and 11-12, training on `COCO+TextCaps' leads to worse performance than only using COCO/TextCaps (rows 1,4,7,10). It means that simply improving the sampling ratio of these two datasets can not handle the domain shift problem, which is already a quite challenging task. However, combining COCO and TextCaps datasets for training is  more suitable for complex real scenarios. 
In this way, the \rev{well-trained} model is able to `watch' visual objects and `read' texts in images.

\section{More ablation studies about losses}\label{sec:loss}

\begin{figure}[t]
\renewcommand\thefigure{S2}
\centering
\begin{minipage}[t]{0.45\textwidth}
\centering
    \includegraphics[width=0.9\textwidth,height=0.6\textwidth]{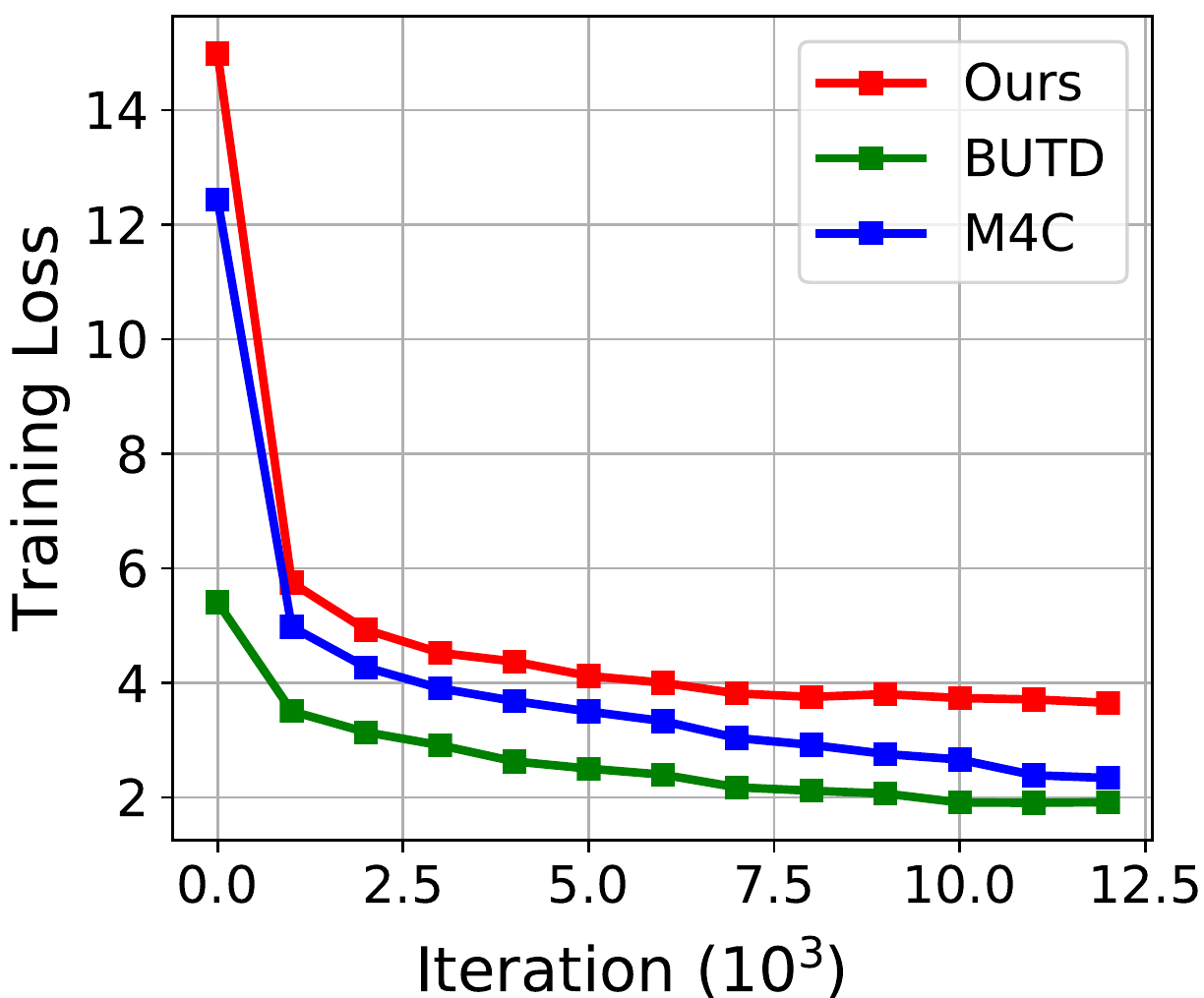}
\end{minipage}
\hspace{0.28in}
\begin{minipage}[t]{0.45\textwidth}
\centering
    \includegraphics[width=0.9\textwidth,height=0.6\textwidth]{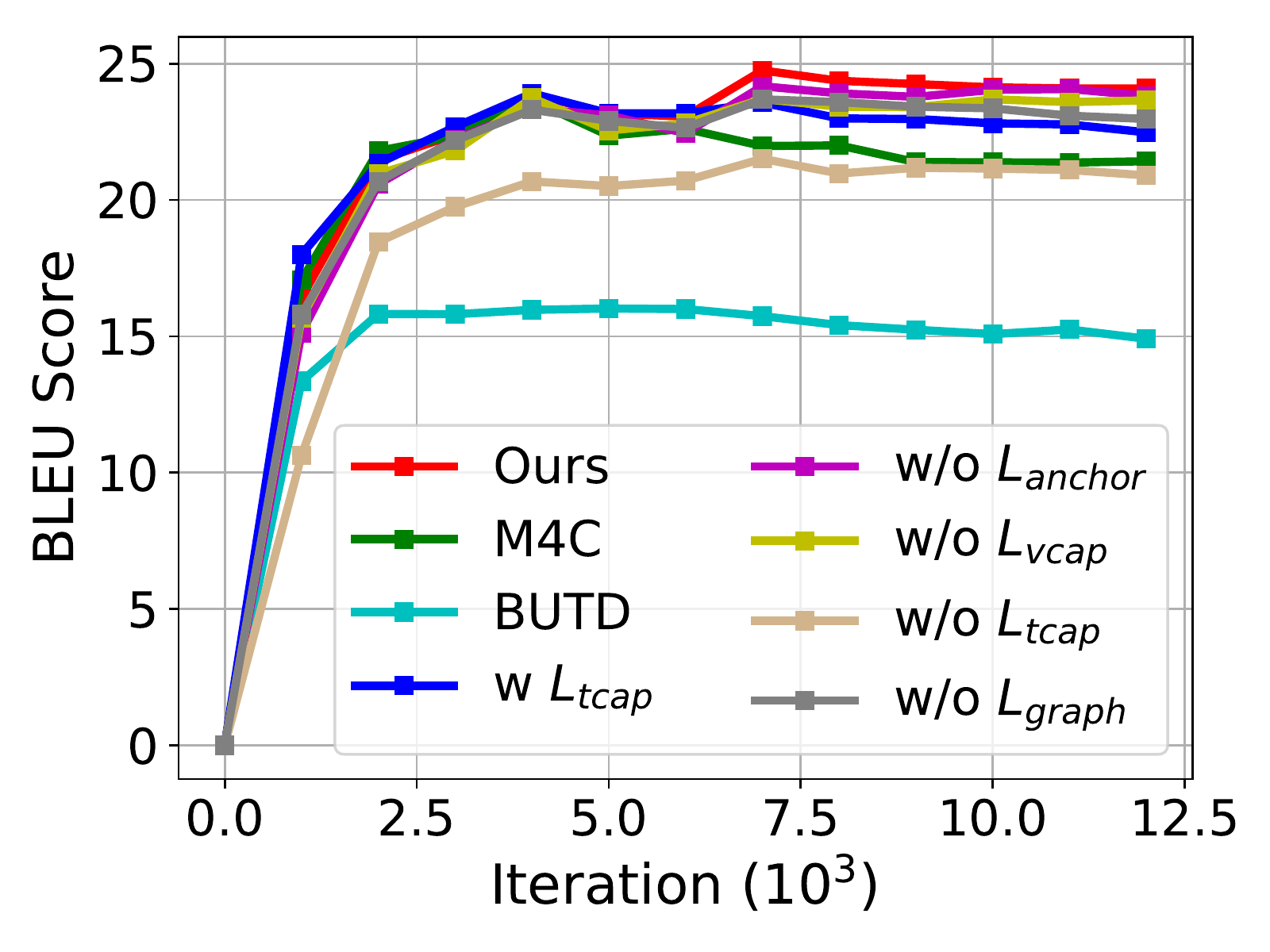}
\end{minipage}
\caption{\textbf{Left}: The training loss of different methods on the TextCaps training set. \textbf{Right}: The BLEU scores under different iterations on the TextCaps validation set. The `M4C' denotes M4C-Captioner model. The `w/o' denotes `Ours' without a specific loss term. For instance, `w/o $\mL_{tcap}$' denotes our method without $\mL_{tcap}$ while `w $\mL_{tcap}$' denotes our method only using $\mL_{tcap}$.}
\label{fig:my_loss}
\end{figure}

As shown in Figure~\ref{fig:my_loss}, we also compare our method with baselines in terms of the training loss and the BLEU score. Since our total training loss contains four terms \rev{$\mL_{\{anchor, graph, vcap, tcp\}}$}, the value of our training loss is little higher than the compared methods. Our method tends to converge after 7k iteration and achieves the highest BLEU score on the validation set. Compared with the considered methods, our method has better generalisation ability to overcome the overfitting problem. \rev{To further measure the importance of losses in our method, we conduct several ablation studies, such as removing each loss term. In particular, we can train our model in an end-to-end miner (only using $\mL_{tcap}$). From the results, the importance of the losses can be formulated as: $\mmL_{tcap}\small{>}\mmL_{vcap}\small{>}\mmL_{graph}\small{>}\mmL_{anchor}$.}

\section{More visualisation analysis}\label{visualisation}
\begin{figure*}[t]
\renewcommand\thefigure{S3}
    \centering
    \includegraphics[width=\textwidth]{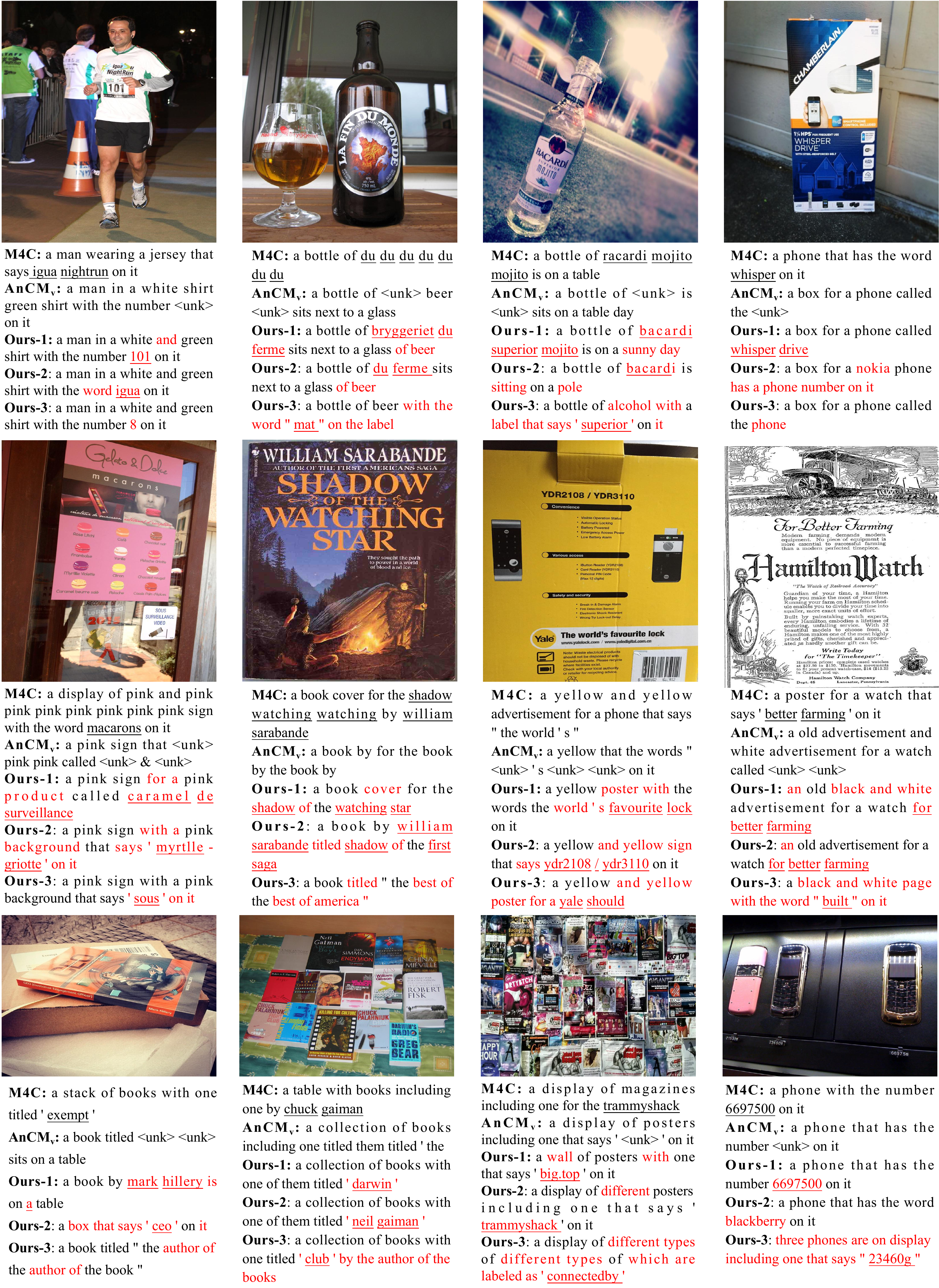}
    \caption{More visualisation results on the TextCaps validation set. For better visualisation, the underlined word is copy from OCR tokens. The modified tokens are viewed in \textcolor{red}{red} colour.}
    \label{fig:s1}
\end{figure*}
\begin{figure*}[t]
\renewcommand\thefigure{S4}
    \centering
    \includegraphics[width=\textwidth,height=0.75\textwidth]{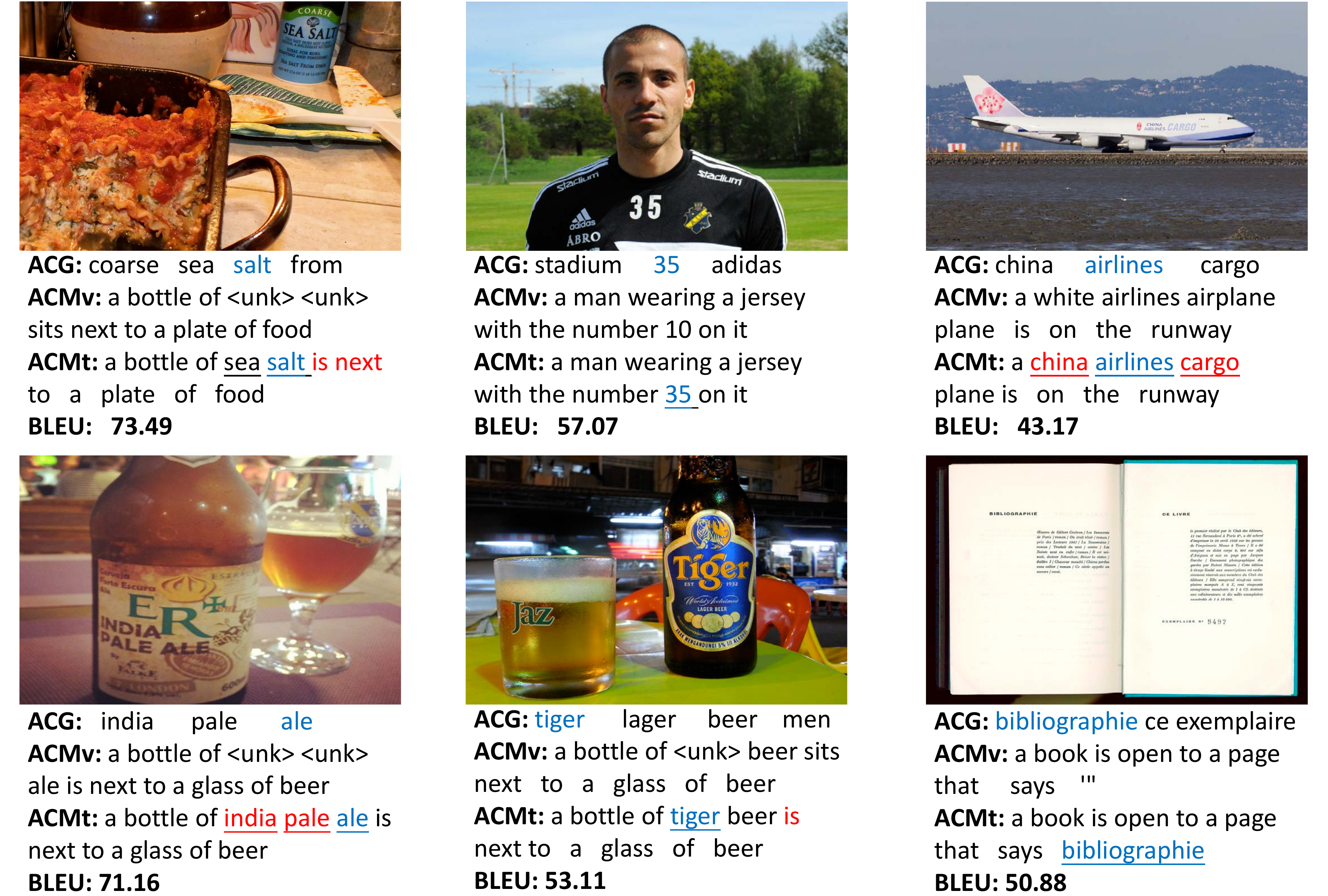}
    \caption{Visualisation results on controllability of our method. For each image, we show the top-1 anchor-centred graph (ACG) and the generated captions of visual-captioner (AnCM$_v$) and text-captioner (AnCM$_t$). In particular, we report the BLEU score of text-captioner's output. For better visualisation, the anchor in ACG is viewed in \textcolor{cyan}{blue} colour, the underlined word is copy from ACG and the modified tokens are viewed in \textcolor{red}{red} colour.}
    \label{fig:acg_control}
\end{figure*}
\begin{figure*}[t]
\renewcommand\thefigure{S5}
    \centering
    \includegraphics[width=\textwidth]{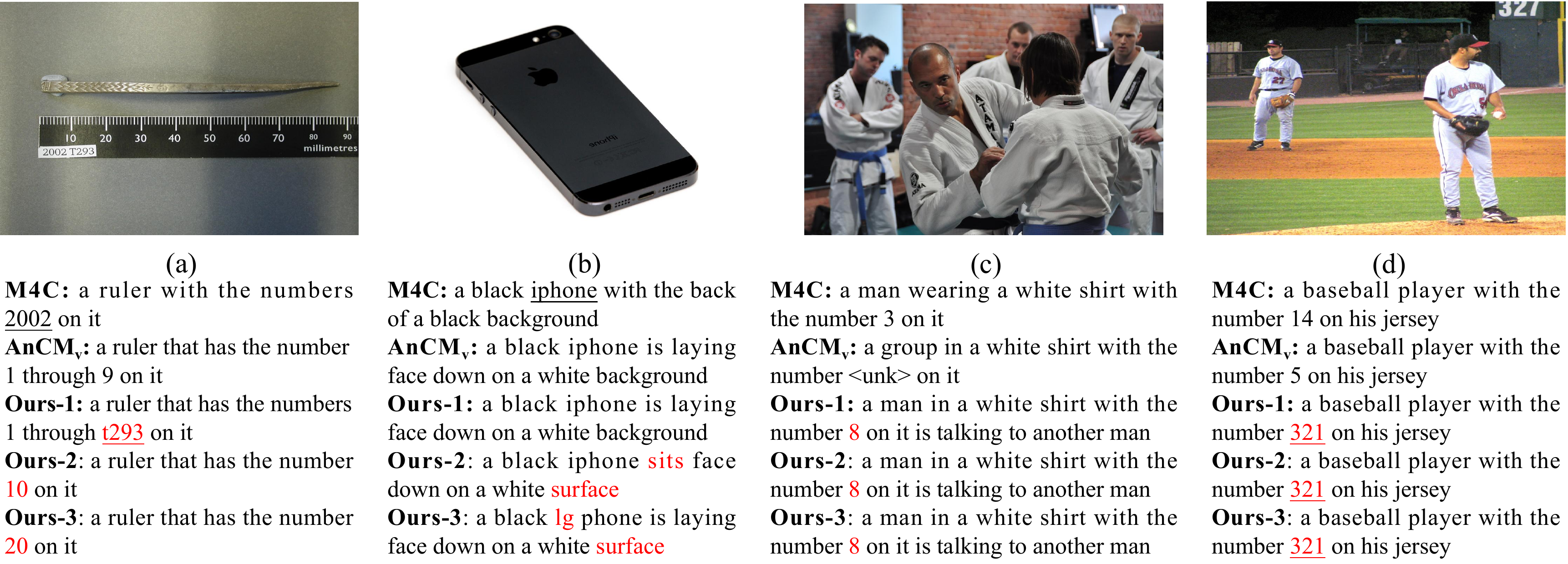}
    \caption{Some failure cases of our model on the TextCaps validation set. The $<$unk$>$ denotes `unkown' token. The underlined word is copy from OCR tokens. The modified tokens are viewed in \textcolor{red}{red} colour.}
    \label{fig:s2}
\end{figure*}

\rev{In this section, to further measure the qualities of our method's generation, we provide more visualisation results on TextCaps validation set. Specifically, we fist show more successful results of our method in Figure~\ref{fig:s1}. Then, demonstrate the controllability of our method in Figure~\ref{fig:acg_control}. Last but not least, in Figure~\ref{fig:s2}, we provide some typical failure cases to evaluate our method more objectively.}

\rev{As shown in Figure~\ref{fig:s1}, compared with M4C-Captioner, our method is able to describe images from different views and cover more OCR tokens, represented as `Ours-*'. In particular, our proposed AnCM is a progressive captioning module that AnCM$_v$ first adopts visual information to generate a global caption and then AnCM$_t$ refines the caption based on the text information of ACGs. Note that, the refining process not only to simply replace $<$unk$>$ token but also to revise the entire caption in terms of language syntax and contents. In addition, extensive experiments in the main paper also demonstrate the effectiveness of our method.}

\rev{Based on anchor-centred graphs (ACGs), our method is able to generate multiple controllable image captions. To demonstrate the controllability of our method, we provide more visualisation about the generated captions aligned with the ACGs . As shown in Figure~\ref{fig:acg_control}, the generated caption of AnCM$_t$ is aligned with the ACGs. We also can see that the generated captions always contain anchors. One possible reason is that our model takes the most important OCR token as an anchor while the other tokens in ACG are used to aggregate information to the anchor. And thus the generated caption is supposed to be at least anchor-related.}

As shown in Figure~\ref{fig:s2}, we also provide typical failure cases to further analyse the performance of our method. 1) Although some images could be correctly describe via one global caption,
our model still tends to output multiple diverse captions, which might be correct but uninformative, such as `\emph{ruler ... has number 10/20}' in (a).
2) Due to dataset bias, the model tends to generate words with high frequency in training set, such as \emph{brand name is iphone or lg} in (b). 3) More critical, our model is sensitive to anchor-centred graphs (ACGs). As shown in (c)-(d), if the OCR recognise system fails to detect or only detects few OCR tokens in an image, our model will be degraded to existing models that only generate a global caption since we have trouble contracting different ACGs. 



\end{document}